\documentclass[journal,twoside,web]{ieeecolor}

\usepackage{makecell}
\usepackage{generic}
\usepackage{cite}
\usepackage{amsmath,amssymb,amsfonts}
\usepackage{algorithm}
\usepackage{algorithmicx}
\usepackage{xcolor}
\usepackage{hyperref}
\usepackage{algpseudocode}
\usepackage{graphicx}
\usepackage{booktabs}
\usepackage{textcomp}
\usepackage[final]{changes}
\def\BibTeX{{\rm B\kern-.05em{\sc i\kern-.025em b}\kern-.08em
    T\kern-.1667em\lower.7ex\hbox{E}\kern-.125emX}}
\begin{document}
\title{Medical-Checklist: Assessing the Comprehension of Medical Images by Multimodal Models}
\author{Bannapol Limanond, Masanori Suganuma, and Takayuki Okatani, \IEEEmembership{Member, IEEE}
\thanks{Manuscript received \today. This work was supported by JST SPRING, Grant Number JPMJSP2114.} 
\thanks{\copyright{} 2026 IEEE.  Personal use of this material is permitted.  Permission from IEEE must be obtained for all other uses, in any current or future media, including reprinting/republishing this material for advertising or promotional purposes, creating new collective works, for resale or redistribution to servers or lists, or reuse of any copyrighted component of this work in other works.}}

\maketitle
\begin{abstract}
This paper introduces a new benchmark test, Medical-Checklist, for assessing medical multimodal models. The recent advancements in multimodal models have demonstrated significant potential in the field of medical vision-language tasks. However, it is becoming increasingly clear that evaluating these models' performance, whether they are applied to natural or medical images, is challenging. The critical question is whether the models can accurately understand an input image while associating it with relevant input text. To address this, Medical-Checklist imposes a binary test on the models: they are given an image and two captions, where one is correct and the other incorrect, and the model must select the correct one. The incorrect caption contains a single medical concept (word or phrase) that is inaccurately substituted from the correct caption. Although the task is simple, this simplicity enables the unified assessment of diverse multimodal models designed and learned on different principles. It also enables us to verify whether models correctly understand a wide range of medical concepts across various medical sub-domains. Medical-Checklist is designed to reduce potential biases in data and to enable evaluation of the models' ability to handle out-of-distribution inputs, which were difficult in existing datasets. When evaluating four state-of-the-art medical multimodal models with Medical-Checklist, it was revealed that despite their excellent performance in specific tasks such as Med-VQA, they may not correctly understand images, suggesting a long journey ahead for clinical application. 
The dataset and code will be made public upon acceptance.
\end{abstract}


% ****************** Figure ******************
\begin{figure*}[!t]
    \centering
    \includegraphics[clip, trim=0cm 0cm 0cm 0cm, width=\textwidth]{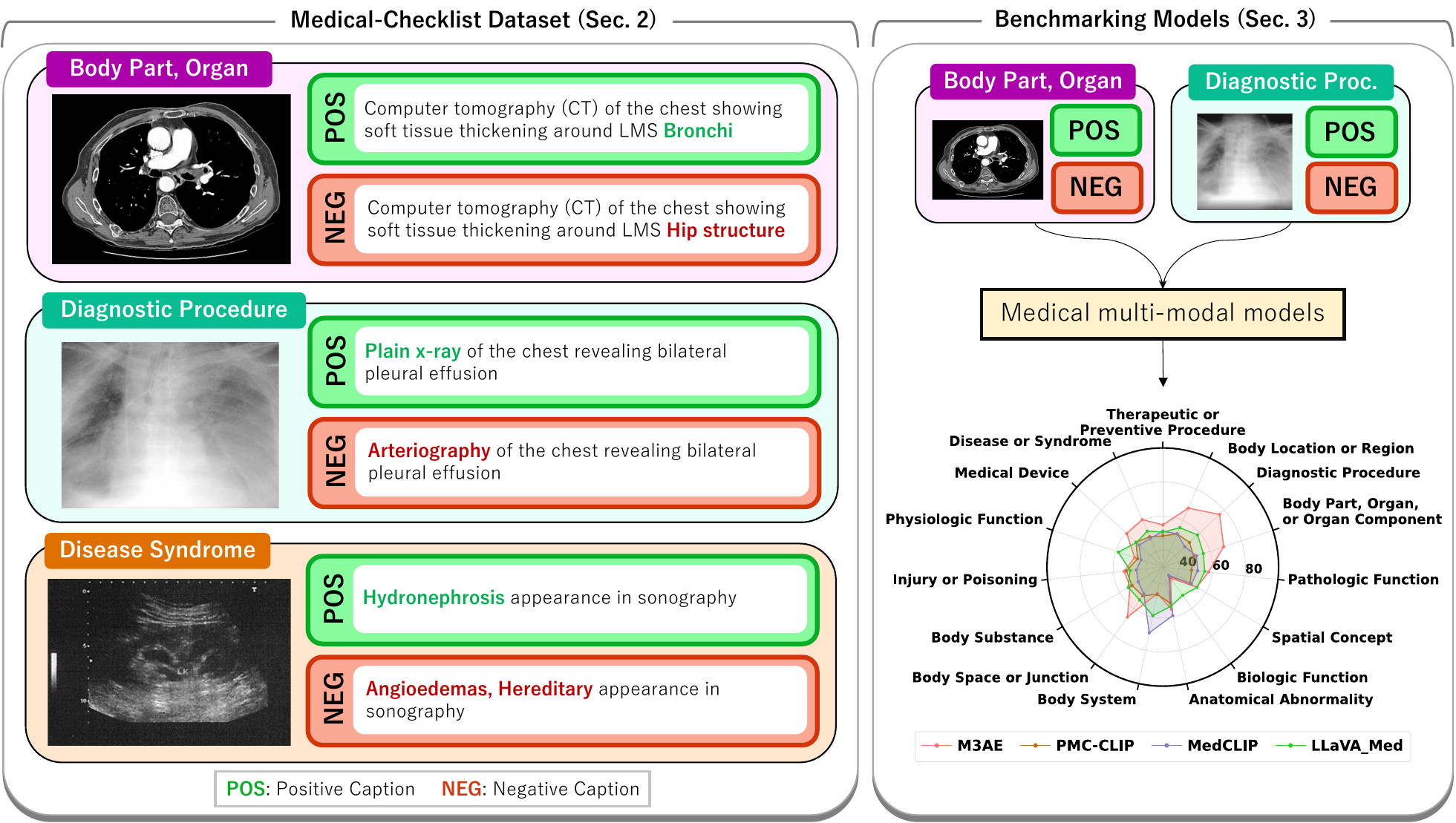}
    \caption{Example tests from Medical-Checklist (left) and the overview of benchmarking medical multimodal models (right). }
    \label{fig:OverviewExample}
\end{figure*}
% ****************** Figure ******************

\section{Introduction}
\label{sec:introduction}
\IEEEPARstart{M}{ultimodal} models have demonstrated remarkable success in recent years, achieving state-of-the-art results in vision-language tasks~\cite{li2021align,li2022blip,yang2022vision}. Recent advancements in the field have led to the development of multimodal large language models (MLLMs), which effectively combine advanced pre-trained large language models (LLMs) with a vision encoder, enabling the transformation of visual inputs into the space of natural language texts \cite{liu2024llava,li2023blip,zhu2023minigpt,radford2021learning}. This integration enables the MLLMs to leverage the unmatched capabilities of LLMs such as robust language generation, zero-shot transfer capabilities, etc.
These advancements have also led to significant progress in the medical field. Multimodal models, trained on various data in the medical domain, have exhibited significantly improved capabilities,
particularly in processing and understanding complex medical images and associated text~\cite{chen2022arl,chen2022m3ae,li2023llavamed,wang2022medclip,lin2023pmcclip}. These models have set new benchmarks in a variety of medical vision-language tasks such as medical visual question answering (Med-VQA), patient report generation, and more \cite{ben2021overview,liu2021slake,zhang2023pmcvqa,zhang2023large,pelka2018roco,wu2021melinda,demner2016preparing,subramanian2020medicat,irvin2019chexpert,he2020pathvqa,lau2018vqarad}.
Nevertheless, we should not uncritically accept the aforementioned achievements. The difficulty in accurately assessing the performance of multimodal models is becoming more apparent in the domain of natural images. Multimodal models, such as CLIP, have demonstrated exceptional performance in tasks like zero-shot image classification \cite{radford2021learning,li2022blip,li2021align}. However, it has been shown that these models only possess a superficial understanding of images \cite{zhao2022vlchecklist,yuksekgonul2023aro,thrush2022winoground}. In the medical sector, there have recently been multiple papers assessing the capabilities of GPT-4V, which is considered to be the most sophisticated MLLM for now, in the context of medical image diagnosis \cite{han2023comparative,buckley2023accuracy,yang2023performance,chen2023enhancing,wu2023gpt4vmed,yan2023gpt4vmedapp}. Yet, the findings of these studies are markedly inconsistent, with some suggesting the model is ready for practical application \cite{han2023comparative,buckley2023accuracy,yang2023performance,chen2023enhancing} and others arguing it is entirely unfit for such purposes \cite{wu2023gpt4vmed,yan2023gpt4vmedapp}. This variation in findings emphasizes the complexity of evaluating multimodal models accurately.
In other words, achieving outstanding results on a specific dataset for a particular task does not automatically validate the model's effectiveness, especially in the real world. The dataset may contain inherent biases, which the model could simply be exploiting. Additionally, the challenge of evaluating models against out-of-distribution (OOD) inputs—inputs that differ from those seen during training—remains. These point to limitations of the conventional evaluation methodology, which involves splitting a closed dataset into training and testing sets. 
In this study, we propose a new benchmark, Medical-Checklist, to evaluate medical multimodal models from a different perspective than prior work. The benchmark consists of a large number of medical images, each paired with two captions: one that correctly describes the image and one that contains a single randomly substituted medical term (a word or phrase), creating an incorrect caption. For example, consider a chest x-ray image paired with contrasting captions: ``Plain x-ray of the chest revealing bilateral pleural effusion'' (correct) versus ``Arteriography of the chest revealing bilateral pleural effusion''(incorrect); see Fig.~\ref{fig:OverviewExample}.

Since the incorrect captions are generated through random substitution, most of them are medically nonsensical. Therefore, a model with correct medical knowledge should be able to select the correct caption.
The research question of this study is whether state-of-the-art multimodal models specialized for medical imaging can actually do so.

This research question distinguishes our work from prior studies on multimodal AI applications to medical images, such as medical VQA. While multiple-choice questions are also common in existing medical VQA datasets, the distractor options are typically clinically plausible. This is natural, as the primary goal of those benchmarks is to assess whether a model can make fine-grained clinical decisions in difficult scenarios. In contrast, the incorrect captions in our benchmark—although grammatically valid due to noun phrase substitution—are clinically implausible. Our benchmark aims to test whether medical multimodal AI models can detect such clearly incorrect answers. 
The motivation is to assess whether these models possess basic medical knowledge and, more broadly, to understand—together with existing benchmark results—what they truly comprehend and what they do not. Accurately identifying what an AI model does and does not understand remains a major challenge across AI research in general.
% The motivation is to verify whether these models possess basic medical knowledge, and more broadly, to better understand what they do and do not actually comprehend.}

Medical-Checklist uses a simple binary-choice task, which offers several advantages.
% \deleted{While binary tests are straightforward, this simplicity facilitates the creation and application of tests for model evaluation. }
First, it allows for an increase in test diversity, enabling a more comprehensive assessment of models' understanding of various medical concepts. Moreover, it aids in evaluating models' generalizability across in- and out-of-distribution inputs. Additionally, this simplicity permits the comparison of diverse multimodal models designed for varied tasks and trained using different methodologies and datasets\footnote{A task that involves selecting the more relevant option aligns well with CLIP's training objective and is thus natural for the models based on CLIP, while it may be slightly disadvantageous for MLLMs. However, considering the level of dialog capability expected of MLLMs, this task is sufficiently primitive, and high performance should reasonably be expected—making this evaluation method appropriate.}. Ultimately, it contributes to mitigating inherent biases, preventing the possibility of models exploiting such biases. For instance, binary tests can be designed to guarantee an equal distribution between the two possible outcomes, ensuring fairness in the evaluation process.

We evaluate four state-of-the-art medical multimodal models~\cite{wang2022medclip,lin2023pmcclip,chen2022m3ae,li2023llavamed}. Although all of them demonstrate strong performance on existing benchmarks, none of them are able to pass the Medical-Checklist: they consistently fail to distinguish clearly incorrect answers from correct ones with the expected level of accuracy. This indicates that despite their success in answering clinically plausible VQA questions, these models lack a fundamental understanding of core medical concepts. Further analysis is presented in the following sections.
% \deleted{We conduct experimental evaluations on four state-of-the-art medical multimodal models~\cite{wang2022medclip,lin2023pmcclip,chen2022m3ae,li2023llavamed}, each developed under different disciplines. Our findings reveal their inability to pass our ``Medical-Checklist,'' highlighting a fundamental lack of understanding of medical concepts.}
%

% ****************** Table ******************
\begin{table*}[!t]
\centering
\caption{Distribution of Medical-Checklist categories and alternative terms}
\label{table:distribution}
\begin{tabular}{lrrl}
\toprule
Category & Count & Percentage & Example of Alternative Terms \\
\midrule
Spatial Concept        & 12,027 & 18.37 & Junctional, Linear, Unifocal, Foraminal, Loop, Proximity, Left, Right, Ovoid shape, Parasternal \\
Pathologic Function    & 3,902  & 5.96 & Stenosis, Atelectasis, Lysis, Microcalcification, Reflux, Microaneurysm, Hypoplasia, Hyperplasia   \\
Body Part or Organ     & 13,851 & 21.16 & Conjunctiva, Permanent permolar tooth, Scaphoid bone, Cerebral Peduncle, Zygomatic bone  \\
Diagnostic Procedure   & 6,354  & 9.71 & Urethrogram, CT, MRI, Autopsy, Cystography, Esophagram, Radioscopy, Plain chest X-ray   \\
Body Location          & 3,584  & 5.47 & T6 level, C3 level, Spleen acupuncture point SP1, Intracranial, Intervertebral, Coronal \\
Therapeutic Procedure  & 6,930  & 10.59 & Balloon Dilatation, Extraction, Amputation, Intramedullary Nailing, Suction drainage, Acid Etching \\
Disease or Syndrome    & 9,332  & 14.26 & Biloma, Cholelithiasis, Pituitary Diseases, Arthropathy, Empyema, Mycetoma, Hematuria \\
Medical Device         & 3,637  & 5.56 & Apron, Blades (device), Prosthesis, Cannula device, transducers, Gastric Balloon   \\
Physiologic Function   & 463   & 0.71 & Electric impedance, Tropism, Blood circulation, Muscle contraction, Excretory Function  \\
Injury or Poisoning    & 1,080  & 1.65 & Retropneumoperitoneum, Tissue damage, Periprosthetic fractures, Microfractures, Incised wound   \\
Body Substance         & 970   & 1.48 & Exudate, Dental cementum, Feces, Dentin, Trichobezoars, Hyaline Substance, Venous plasma   \\
Body Space             & 1,666  & 2.54 & Prepontine Cistern, Pulp Canals, Decussation, Dento-alveolar joint, L5-S1 intervertebral space\\
Body System            & 187   & 0.29 & Peripheral nervous System, Autonomic nervous System, Skeletal system, Genitourinary system   \\
Anatomical Abnormality & 1,250  & 1.91 & Absent left coronary artery, Dilation of ureter, Cortical irregularity, Subperiosteal bone resorption   \\
Biologic Function      & 231   & 0.35 & Bioluminescence, Anabolism, Ingestion, Process of secretion, Corticogenesis, Biological Signaling   \\
\midrule
\textbf{Total}         & \textbf{65,464} & \textbf{100.0} &  \\
\bottomrule
\end{tabular}
\end{table*}
% ****************** Table ******************

%

\section{Related Work}
\label{sec:related-work}
\subsection{Multi-modal Medical Tasks} 
%
%1. add bias explain poteintial of it, because it not binary and have possible answer.
%2.diff model method develop inder different disciplines use diff beanchmark lead to cannot interpret and cannt compare results.
%3.More cited
%
Due to the rapid advancements in medical multimodal models~\cite{alayrac2022medflamingo,li2023llavamed,wu2023generalist}, the field of multimodal medical benchmarks has attracted significant attention in recent times. Notable tasks within this domain include medical image-text classification, medical image-caption retrieval, report generation, biomedical visual chatbot, chest X-ray phrase grounding, and medical visual question answering.
Medical image-text classification~\cite{wu2021melinda} involves the multi-label classification of text describing medical experiment methods, based on their corresponding images.
Medical image-caption retrieval~\cite{pelka2018roco,demner2016preparing} facilitates the process of retrieving captions that correspond to given texts and vice versa. 
Report generation \cite{zhang2023large,subramanian2020medicat,irvin2019chexpert,lin2023pmcclip} encompasses the automated creation of medical reports that summarize findings from images, as well as impressions from physicians.
Biomedical visual chatbot~\cite{li2023llavamed} enables an open-ended interactive dialogue between humans and an AI system, focusing on medical images. 
Chest X-ray phrase grounding~\cite{boecking2022making} involves the model being presented with medical phrases, after which it is required to identify and highlight specific regions within a chest X-ray image.
Medical visual question answering (Med-VQA)~\cite{ben2021overview, liu2021slake,lau2018vqarad,he2020pathvqa,ben2019vqa,zhang2023pmcvqa,hu2024OmniMedVQA,alayrac2022medflamingo,naseem2024kpathvqa} involves models that are tasked with responding to medical questions associated with images. This process requires a sophisticated integration of visual understanding and contextual knowledge to provide accurate answers based on the presented visual data.
However, benchmarks used in existing studies often exhibit limitations. First, these studies often limit their evaluations to results that are not directly comparable, as different works may select various tasks for assessment, thereby complicating comparative analyses. This issue arises due to the incompatibility of certain visual language models with specific tasks. For example, Med-CLIP~\cite{wang2022medclip} cannot be directly fine-tuned for Med-VQA.
Moreover, tasks such as medical image-text classification and Med-VQA are susceptible to inherent biases, primarily because the answer data distribution is not evenly balanced. This imbalance could potentially lead to models exploiting these biases.
Additionally, the robustness of a model against input noise remains uncertain. For example, it is unclear how a model would perform if a medical term within the input were replaced by a random term — a modification that could significantly impact its interpretability and accuracy. 
Ultimately, with the exception of the biomedical visual chatbot task, existing studies limit their evaluations to ``in-distribution'' settings, where models are tested on a separate portion of the same dataset used for training. This approach restricts insight into the models' true generalization capabilities, as it does not assess their performance on entirely new, unseen data.
\subsection{Evaluation of MLLMs in Medical Image Diagnosis} Several preprints of studies have recently been published that test the application of GPT-4V~\cite{gpt4vcontribution} in the medical image diagnosis. Interestingly, they present two divergent perspectives. Several studies advocate for the utility of GPT-4V. This includes the comparative studies of the performance between human physicians and GPT-4V~\cite{han2023comparative,buckley2023accuracy}. One study~\cite{yang2023performance} shows that GPT-4V passed the full USMLE exam\footnote{https://www.usmle.org/}, showing potential for clinical decision support. Another research group~\cite{chen2023enhancing} focuses on prompt engineering with the model, demonstrating remarkable performance improvements. However, other studies draw the exact opposite conclusion. A paper~\cite{wu2023gpt4vmed} presents a qualitative evaluation of GPT-4V on multiple clinical tasks, e.g., disease diagnosis and report generation, raising questions about the model's suitability for clinical tasks. Additional findings~\cite{yan2023gpt4vmedapp} identify seven key limitations of GPT-4V's performance in clinical diagnosis contexts, including an over-reliance on textual information and difficulties in assessing medical object sizes. This conflict highlights the challenge of accurately assessing multi-modal models on medical tasks, which demand a thorough comprehension of medical concepts. Additionally, it underscores the inherent difficulty in guaranteeing no overlap between training and testing data when evaluating proprietary models like GPT-4V, for which the training datasets are not publicly disclosed.
%

% ****************** Figure ******************
\begin{figure*}[!t]
    \centering
    \includegraphics[clip, trim=2cm 4cm 2cm 4cm, width=\textwidth]{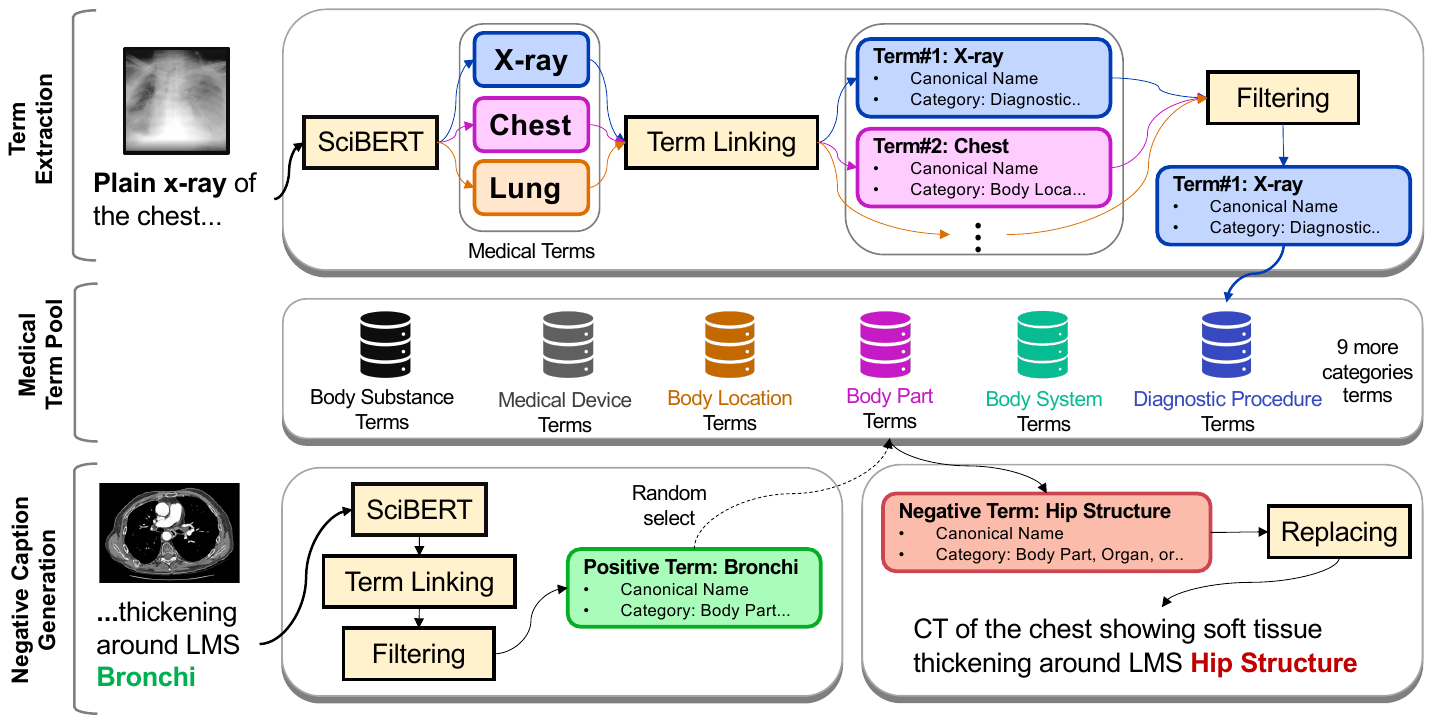}
    \caption{Overview of the dataset creation process.}
    \label{fig:generation}
\end{figure*}
% ****************** Figure ******************

%
\section{Method}
\label{sec:method}
We create a benchmark test, named Medical-CheckList, to evaluate multimodal models' understanding of medical imagery combined with textual descriptions. We adopt a binary classification approach, inspired by studies in natural image understanding \cite{zhao2022vlchecklist,thrush2022winoground,yuksekgonul2023aro}, enabling us to cover a wide array of medical concepts across various subdomains. It requires the models to choose between two captions: one accurately describes the image, and the other, referred to as a negative caption, includes a subtle yet critical modification—a change of a single medical term (e.g., changing `Bronchi' to `Hip structure'), as shown in Fig.~\ref{fig:OverviewExample}.
\subsection{Dataset Specification}
\begin{figure*}[!t]
    \centering
    \includegraphics[clip, trim=1cm 3cm 1cm 3cm, width=\textwidth]{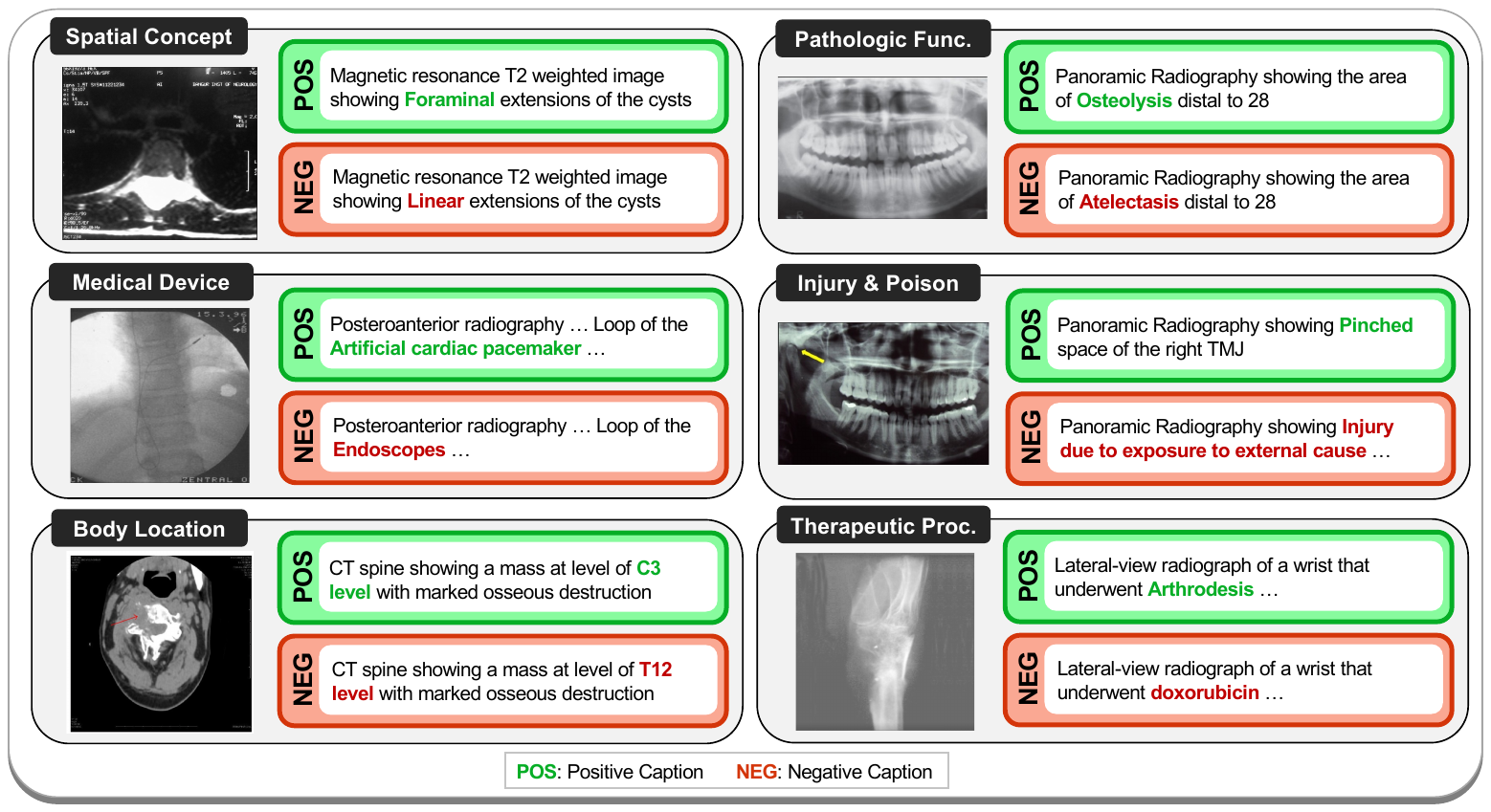}
    \caption{More qualitative examples from the Medical-Checklist.}
    \label{fig:MoreExample}
\end{figure*}

The Medical-Checklist contains 65,464 binary tests classified by 15 diverse categories of medical concepts,
such as `Disease or Syndrome' and `Body Part, Organ, or Organ Component.'
It is built upon 53,556 images, sourced from the MedICat \cite{subramanian2020medicat} and ROCO datasets \cite{pelka2018roco}. The two datasets contain a variety of medical imaging modalities, such as radiology, histology, and scope procedures, and collectively contain over 300,000 medical image-text pairs. 
Each test entry includes: 1) an input image; 2) a positive caption, which is a caption associated with the image and sourced from the original dataset; 3) a negative caption, generated by altering a single medical term (word or phrase) in the positive caption; and 4) the category of both the replaced and the replacing terms, out of 15 pre-defined categories, ensuring both belong to the same category. Overall, there are 6,293 unique medical terms involved in these replacements; see Table~\ref{table:distribution}. Additionally, Fig.~\ref{fig:MoreExample} provides further examples from the Medical-Checklist dataset.
\subsection{Design of the Dataset}
\subsubsection{Desirable Properties}
To achieve the stated goal, we aim for the benchmark to meet three criteria. First, it should evaluate models' resilience across a continuous spectrum of input distributions, ranging from in-distribution to out-of-distribution. For example, models should maintain robustness against input anomalies, like mismatches between an image and its caption.
Second, the benchmark should minimize statistical biases to ensure fair and accurate model evaluations. An example of such bias is when questions starting with ``What is'' consistently favor A in a binary test between A and B. Additionally, the distribution of answers within the entire dataset may disproportionately favor A over B. Notably, many existing VQA-based benchmarks in the medical field do not meet both of these criteria.
Third, the benchmark should facilitate the comparison of results across all types of medical multimodal models, including those trained with image-text contrastive loss, image-text matching loss, and multimodal large language models.
The Medical-Checklist is designed to meet the above criteria as much as possible. It covers a diverse range of captions across various domains. The task of differentiating between the correct and similar negative counterpart enables us to assess models' robustness against input anomalies. Moreover, it adjusts the answers for binary classification to be evenly distributed at 50\% each, thereby mitigating inherent bias.
\subsubsection{Choosing Categories of Medical Concepts}
\label{sec:categories}
To facilitate the evaluation of models tailored to various medical subdomains, we selected 15 categories of medical concepts. These categories are chosen based on the Q\&A pairs found in existing multimodal Med-VQA benchmarks. To ensure consistent use of terminology, we employ the Unified Medical Language System (UMLS)~\cite{bodenreider2004umls}. The 15 categories are: `Anatomical Abnormality,' `Body Part, Organ, or Organ Component,' `Medical Device,' `Body Substance,' `Body System,' `Spatial Concept,' `Body Space or Junction,' `Body Location or Region,' `Diagnostic Procedure,' `Therapeutic or Preventive Procedure,' `Biologic Function,' `Physiologic Function,' `Pathologic Function,' `Disease or Syndrome,' and `Injury or Poisoning.'
Detailed definitions of each concept and its clinical relevance can be found in the supplementary material and on the official UMLS website.\footnote{https://www.nlm.nih.gov/research/umls/index.html}

\subsubsection{Process of Dataset Creation}
We develop the dataset in two steps; see Fig. \ref{fig:generation}. In the first step, we extract medical terms from the captions of the two mentioned datasets, selecting only those that fit into the 15 concept categories. Specifically, we employ SciBERT \cite{beltagy2019scibERT} to identify medical terms in the captions, noting that a single caption may contain several terms. Following this, we assign a canonical name and category to each identified term by referencing the UMLS database, utilizing a string overlap search method based on character 3-grams. Terms that do not match any of the 15 predefined categories are discarded. Finally, we sort and categorize the remaining terms according to their medical categories, creating a term pool that allows structured access in the subsequent step. 
%

% ****************** Figure ******************
% ****************** Figure ******************

%
In the second step, we create negative captions for the image-caption pairs. We start with each image-caption pair by identifying a medical term from the term pool within the caption. Next, we randomly select an alternative term from the same medical category in the pool and replace the original term with this selected one. The negative captions are intentionally crafted to include errors that should be identifiable by a model trained to understand medical images and terminology, regardless of their medical plausibility. It is important to note that a caption may contain multiple terms, leading to the generation of an equal number of negative captions. This process results in the creation of tuples, each consisting of an image, the original positive caption, and a negative caption, together forming a single binary test. Throughout this process, we use SciSpacy~\cite{neumann2019scispacy} to handle text analysis.
\subsubsection{Quality Control and Dataset Validation}
Our creation process follows two UMLS database criteria to ensure quality of the benchmark: Concept IDs (CUIs) and Semantic Types (TUIs). The CUI ensures that negative terms are not synonyms or semantic variants of positive terms. For example, the process cannot swap ``heart attack" with ``myocardial infarction" because both terms map to the same CUI. Similarly, the TUI ensures that positive and negative terms belong to the same medical category. These constraints prevent systematic errors and noise in our benchmark.

We also conducted manual validation to verify the reliability of our creation process. We randomly sampled 100 cases and manually verified: (1) whether the extracted terms were medically relevant, and (2) whether the terms were accurately linked to their corresponding CUIs and TUIs. We observed zero errors in both term extraction and semantic linking. 
\subsection{Benchmarking Multimodal Models}
To assess multimodal models, we assign them the task of selecting the correct caption from a tuple consisting of an image, a positive caption, and a negative caption. As it is a binary classification task, we use average classification accuracy as the evaluation metric, with a chance rate of 50\%. We further assess model confidence and calibration via cross-entropy loss and the Brier score.
There are various multimodal models, each designed and trained differently, necessitating careful consideration for their evaluation. In this study, we evaluate three types of models. The first category includes models trained using image-text contrastive loss, such as Med-CLIP \cite{wang2022medclip} and PMC-CLIP \cite{lin2023pmcclip}. For these models, we leverage the image-text contrastive loss ~\cite{zhao2022vlchecklist,yuksekgonul2023aro} to choose between two input captions based on the higher similarity score, which the model interprets as the correct answer. The second category comprises models trained with image-text matching loss (excluding image-text contrastive loss) like M3AE \cite{chen2022m3ae}. Here, we use the image-text matching loss to determine the correct caption. The third category involves MLLMs, such as Med-LLaVA \cite{li2023llavamed}. MLLMs are typically trained on next-token-prediction tasks using plain image-caption pairs or visual instruction tuning data and do not directly provide a similarity score between an image and a caption. To facilitate selection among two captions, we use a prompting method \cite{gordon2023mismatch}, which effectively guides the model in making a choice. The prompt is as follows:
%

% ****************** algorithmic ******************
\begin{algorithm}[H]
\caption{Prompt for evaluating MLLMs.}
\label{alg:prompt}
\begin{algorithmic}
\State
\State Given the following image and two captions, which caption more accurately describes the image? Please answer with either `Caption A' or `Caption B' only.
\State
\State The image is displayed here: ⟨image⟩.
\State Caption A: ⟨original caption⟩.
\State Caption B: ⟨negative caption⟩.
\end{algorithmic}
\end{algorithm}
% ****************** algorithmic ******************

%
\noindent
We randomize which of Captions A and B contain the positive caption to prevent bias. It is important to note that we reviewed the prompts used in related studies and selected the most effective one for comparison across all models. While this method may not be perfect, similar challenges are encountered in any research involving MLLMs. Furthermore, if a model fails to respond accurately to a plain, standard prompt, this failure should be considered a limitation of the model.
\section{Experimental Results}
\label{sec:experiment}
We conducted experiments to evaluate existing medical multimodal models using the Medical-Checklist.
\subsection{Evaluated Models}
We evaluate the four models: Med-CLIP \cite{wang2022medclip}, PMC-CLIP \cite{lin2023pmcclip}, M3AE \cite{chen2022m3ae}, and LLaVA-Med \cite{li2023llavamed}. 
We use the models provided in their respective official Github repositories, along with their code where fine-tuning was necessary.
It should be noted that during tests of the LLaVA-Med model on Medical-Checklist using the specified prompting method, it often does not follow the instructions correctly. Specifically, it produces outputs in formats other than the intended `Caption A' or `Caption B.' As a result, we have excluded the test entries where it fails—accounting for 51.6\% of all tests—and only calculated the average accuracy using the remaining entries (48.4\%) for its evaluation. The other three models do not have this issue and are evaluated across the entire Medical-Checklist.
To deal with this issue with LLaVA-Med, we conduct an additional experiment where we fine-tune the LLaVA-Med model with instruction tuning, before the binary test. We select `Therapeutic Procedure' from among the 15 medical concept categories, and perform the instruction tuning using the test entries related to this category. By employing this method, the model followed the instructions correctly, successfully identifying all of the entities involved (100.0\%) during its evaluation. Subsequently, we assess the performance of the modified model across tests covering the other 14 concepts. In this experiment with LLaVA-Med, we additionally test more variants of the LLaVA-Med models. Specifically, we fine-tune the original LLaVA-Med model on two tasks simultaneously: instruction following and either PathVQA~\cite{he2020pathvqa} or SLAKE~\cite{liu2021slake}, ensuring that there is no forgetting of either task. For fine-tuning on the Med-VQA tasks, we adhere to the method outlined in \cite{li2023llavamed}. The motivation is to assess the impact of fine-tuning on Med-VQA datasets on the model's performance on the Medical-Checklist. Our results confirm that this simultaneous training does not negatively affect the model's performance on the Med-VQA tasks.
%

% ****************** Table ******************
\begin{table}[ht]
\centering
\caption{Accuracy, Brier score and cross-entropy loss of four medical multimodal models on Medical-Checklist}
\begin{tabular}{lccc}
\toprule
Model      & Accuracy   & Brier score & Log loss \\
\midrule
Random     & 50.00       & 0.2500      & 0.6931   \\
\midrule
MedCLIP    & 50.12       & 0.2508      & 0.6932   \\
PMC-CLIP   & 49.72       & 0.3336      & 1.0860   \\
M3AE       & 61.76       & 0.2227      & 0.6496   \\
LLaVA-Med  & 54.25       & 0.2669      & 0.7440   \\
\bottomrule
\end{tabular}
\label{table:medcheck_perf}
\end{table}
% ****************** Table ******************

% ****************** Table ******************
\begin{table}[ht]
\centering
\caption{Per-category accuracy of the four medical multimodal models on the Medical-Checklist}
\label{table:medchecklist_overall}
\fontsize{7.3}{9}\selectfont
\begin{tabular}{lcccc}
\toprule
Models & MedCLIP & PMCCLIP & M3AE & LLaVA-Med \\
\midrule
% Accuracy & 50.11 & 49.72 & 61.76 & 54.25  \\ 
% \midrule
Spatial Concept        & 49.96 & 49.11 & 53.72 & 53.91 \\
Pathologic Function    & 51.20 & 47.43 & 57.81 & 55.49 \\
Body Part or Organ     & 51.41 & 49.68 & 68.70 & 55.79 \\
Diagnostic Procedure   & 47.92 & 51.65 & 76.22 & 58.25 \\
Body Location          & 51.42 & 51.31 & 67.91 & 55.31 \\
Therapeutic Procedure  & 51.05 & 48.28 & 54.83 & 50.53 \\
Disease or Syndrome    & 48.48 & 49.48 & 60.51 & 53.13 \\
Medical Device         & 49.32 & 51.85 & 59.36 & 51.51 \\
Physiologic Function   & 43.84 & 46.00 & 47.94 & 58.26 \\
Injury or Poisoning    & 46.11 & 52.12 & 53.51 & 49.84 \\
Body Substance         & 47.21 & 51.54 & 52.88 & 54.10 \\
Body Space             & 49.51 & 50.78 & 66.38 & 53.86 \\
Body System            & 69.51 & 46.52 & 45.98 & 59.21 \\
Anatomical Abnormality & 59.12 & 51.52 & 55.44 & 53.52 \\
Biologic Function      & 35.93 & 36.79 & 38.09 & 50.51 \\
\bottomrule
\end{tabular}
\end{table}
% ****************** Table ******************

% % ****************** Figure ******************
% \begin{figure}[t]
%     \centering
%     \includegraphics[clip, trim=0cm 0cm 0cm 0cm, width=\columnwidth]{figures/liman4.png}
%     \caption{An illustration highlighting key terms in the Medical-Checklist.}
%     \label{fig:wordcloud}
% \end{figure}
% % ****************** Figure ******************

% ****************** Table ******************
\begin{table}[t]
\centering
\caption{Performance of instruction-following fine-tuned LLaVA-Med variants on the Medical-Checklist}
\label{table:medchecklist_lvlm}
\begin{tabular}{lccc}
\toprule
Model & \makecell{LLaVA-Med+} & \makecell{LLaVA-Med\\SLAKE+} & \makecell{LLaVA-Med\\PathVQA+} \\
\midrule
Average & 65.07 & 60.18 & 69.72 \\ 
\midrule
Spatial Concept & 69.63 & 61.00 & 69.41 \\
Pathologic Function & 61.28 & 59.87 & 63.17 \\
Body Part or Organ & 65.96 & 64.84 & 74.09 \\
Diagnostic Procedure & 63.77 & 59.07 & 69.58 \\
Body Location & 65.04 & 60.38 & 71.09 \\
Disease or Syndrome & 65.26 & 61.04 & 70.14 \\
Medical Device & 66.43 & 50.45 & 67.42 \\
Physiologic Function & 66.09 & 50.54 & 72.14 \\
Injury or Poisoning & 58.06 & 51.11 & 67.96 \\
Body Substance & 63.81 & 58.56 & 68.97 \\
Body Space & 52.76 & 52.22 & 59.72 \\
Body System & 45.99 & 56.15 & 63.64 \\
Anatomical Abnormality & 47.92 & 54.72 & 59.12 \\
Biologic Function & 78.35 & 48.05 & 77.49 \\
\bottomrule
\end{tabular}
\end{table}
% ****************** Table ******************

%
\subsection{Results of the Four Models}
\label{sec:evaluation_overall}
Table~\ref{table:medcheck_perf} presents the results from the initial experiments evaluating four multimodal models. A major observation is that all four models exhibit relatively poor performance, with accuracy ranging from 49\% to 61\%—barely above the chance rate of 50\%. This suggests that despite their effectiveness in Med-VQA tasks, these models struggle to understand images in conjunction with their paired captions genuinely.

MedCLIP performs at chance on both accuracy and cross‐entropy loss, confirming a near–random guessing strategy. Its only above-chance performance occurs in the body system category, which depends on coarse, image-wide features. In contrast, the other categories are fine-grained and require precise region-to-term alignment that current CLIP-based models fail to achieve.

PMC-CLIP achieved the highest accuracy in the Body System category. It records the highest cross-entropy loss and Brier score, indicating it makes incorrect predictions with confidence and is poorly calibrated. This outcome aligns with previous studies on multimodal models for natural image understanding, which have found that CLIP models often fail to capture the fine-grained relationships between specific image regions and the corresponding words or phrases in captions \cite{zhong2022regionclip}.

In contrast, M3AE achieves the highest accuracy of all evaluated models, although its performance remains low at 61\%. This limitation reflects the model’s reliance on an image–text matching objective. However, the performance gain over other models stems from the use of masked image modeling and masked language modeling objectives. These results highlight the need for training objectives that enable models to distinguish fine-grained pathological anomalies in medical images and texts.

Surprisingly, the results show that LLaVA-Med is overconfident and poorly calibrated. This result is inconsistent with its high performance in the standard evaluations. Although this setup may not fully reflect the optimal use of MLLMs—each of which may favor specific Q\&A formats—it is important to note that the ability to engage in diverse interactions about image content is a fundamental expectation for MLLMs.

A potential concern is whether our benchmark can be solved through text-only reasoning since most negative captions are medically nonsensical. However, this concern does not hold for two reasons. First, the architecture of CLIP-based models and M3AE processes each caption independently---these models have no mechanism to directly compare positive and negative captions to detect textual absurdity. Second, if the task were solvable via text-only reasoning, we would expect MLLMs to achieve near-perfect accuracy. Instead, LLaVA-Med achieves only 54.25\% accuracy. The consistent failures across all four evaluated models confirm that our benchmark cannot be solved through text-only reasoning and requires cross-modal understanding.

% ****************** Table ******************
\begin{table*}[t]
\centering
\caption{Top-performing medical terms and their associated accuracy scores for each model.}
\label{table:top_terms}
\begin{tabular}{ll}
\toprule
\textbf{Model} & \textbf{Top-Performing Medical Terms (Accuracy)} \\
\midrule
MedCLIP & ExAblate 4000 Type 2.0 (0.8000), Ascites (0.6613), Bronchiectasis (0.6591), macular (0.6313) \\
PMC-CLIP & Dental caries (0.6889), macular (0.6768), ExAblate 4000 Type 2.0 (0.6300), Cerebrovascular accident (0.6215) \\
M3AE & Hydrocephalus (0.7697), Subarachnoid Space (0.7677), Kidney (0.7458), Sinogram (0.7400) \\
LLaVA-Med & Kidney (0.5876), Hypostasis (0.5500), Cerebrovascular accident (0.5311), Dental caries (0.5278) \\
\bottomrule
\end{tabular}
\end{table*}
% ****************** Table ******************

%
\subsection{Results of LLaVA-Med Variants}
\label{sec:evaluation_mllm}
To investigate whether LLaVA-Med's mediocre performance is due to its failure to follow instructions, we conduct a second experiment by applying task instruction tuning before testing. We observe a substantial improvement in instruction following: the instruction-tuned LLaVA-Med models achieve a 100\% success rate, significantly outperforming the original model's 48.4\% success rate.
Table~\ref{table:medchecklist_lvlm} shows their performance on the Medical-Checklist. We evaluate three variations: the LLaVA-Med model fine-tuned only on the instruction-following task, denoted as LLaVA-Med+, and two versions of LLaVA-Med fine-tuned on SLAKE and PathVQA, as well as instruction tuning, denoted as `LLaVA-Med-SLAKE+' and `LLaVA-Med-PathVQA+,' respectively.
Despite the successful instruction tuning, LLaVA-Med+ shows only a modest improvement in accuracy, rising to 65.07\% from 59.52\%. The highest performance is achieved by LLaVA-Med-PathVQA+, with an accuracy of 69.72\%. Yet, this is still considered low, underscoring the earlier conclusion that the models lack a true understanding of the input images and captions. The slight advantage of LLaVA-Med-PathVQA+ over LLaVA-Med+ suggests that fine-tuning on PathVQA may enhance the model's image comprehension abilities. However, the poorer performance of LLaVA-Med-SLAKE+, compared to LLaVA-Med+, challenges the hypothesis that fine-tuning on a VQA task universally improves model capabilities. The difference between LLaVA-Med-SLAKE+ and LLaVA-Med-PathVQA+ could be attributed to differences in image modalities between PathVQA, which focuses on pathology images, and SLAKE, which is centered on radiology images; this may align better with Medical-Checklist. 
\subsection{Analysis of Factors Influencing Model Performance}
We conducted experiments to explore various factors influencing model performance on the Medical-Checklist dataset. These experiments aim to understand the impact of different input characteristics and model-specific behaviors on accuracy, including caption length, vocabulary preferences, and the similarity of replacement terms.
\subsubsection{Caption Length}
\begin{figure}[ht]
    \centering
    \includegraphics[clip, trim=0cm 0cm 0cm 0cm, width=\columnwidth]{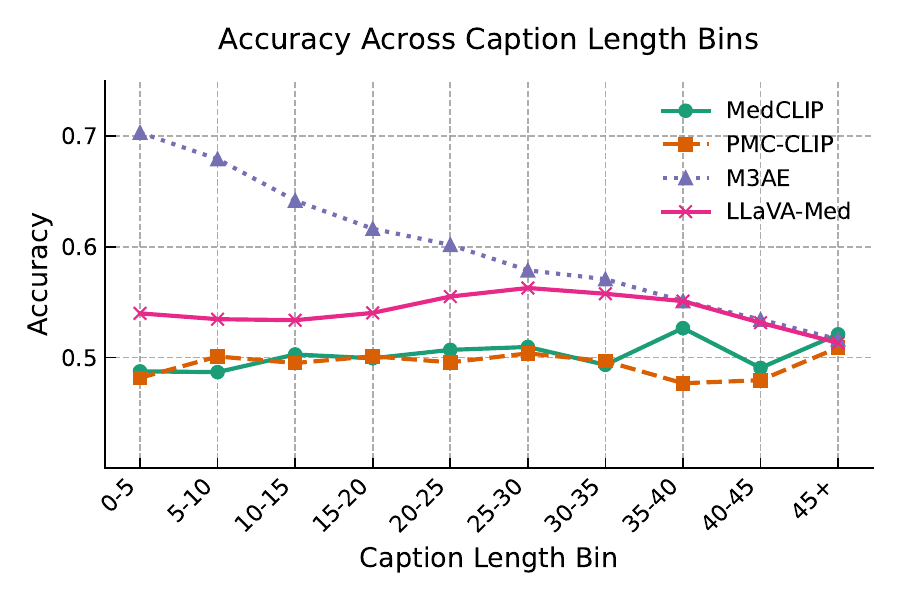}
    \caption{Model accuracy on Medical-Checklist across caption length.}
    \label{fig:acc_length}
\end{figure}

To investigate how the length of captions affects model accuracy, we grouped the captions into nine intervals based on the token count. The performance of each model was analyzed across these intervals. Fig.~\ref{fig:acc_length} shows the accuracy of each model within these bins. The results show that MedCLIP, PMC-CLIP, and LLaVA-Med maintain relatively 50\% accuracy in all bins, indicating that their performance is not affected by caption length. In contrast, M3AE achieves higher accuracy in shorter captions. We attribute this to M3AE’s token-level supervision via Masked Image Modeling (MIM) and Masked Language Modeling (MLM). With shorter captions, each masked token represented a larger proportion of the input, providing a stronger learning signal. 

\subsubsection{Vocabulary Preference}
This experiment explores the vocabulary biases of each model by analyzing the set of medical terms each model prefers. Table~\ref{table:top_terms} shows the vocabulary preferences of each model. The results show that each model prefers a different set of medical terms. It also suggests that any vocabulary bias in the Medical-Checklist dataset is minor or insufficient to impact model performance.
\subsubsection{Replacement-Term Similarity}
\begin{figure}[ht]
    \centering
    \includegraphics[clip, trim=0cm 0cm 0cm 0cm, width=\columnwidth]{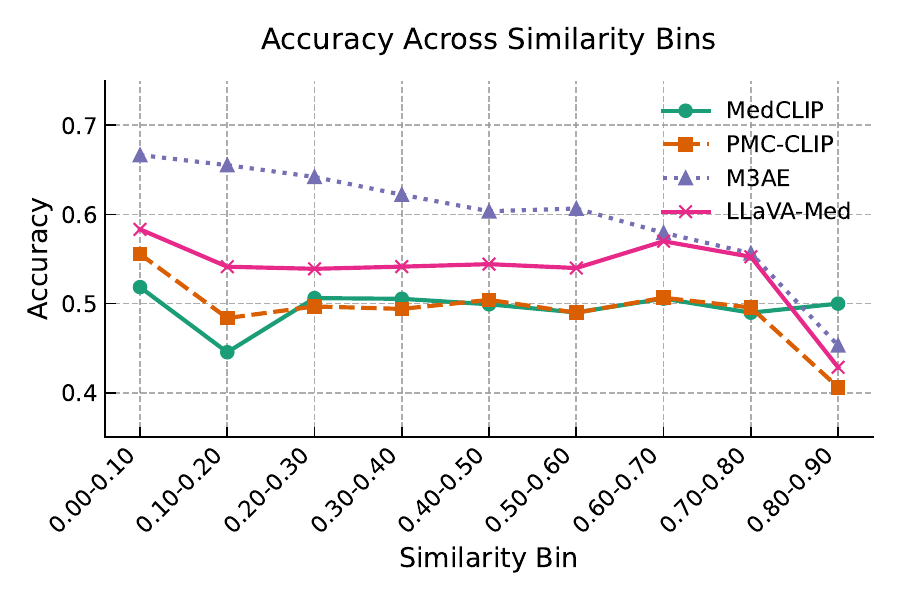}
    \caption{Model accuracy on Medical-Checklist across term similarity.}
    \label{fig:acc_similarity}
\end{figure}

The similarity of replacement terms between positive and negative captions was quantified using SapBERT~\cite{liu2021self} cosine similarity. These scores were divided into nine bins, ranging from 0 to 1, with 0 indicating completely unrelated terms and 1 indicating identical meaning. Fig~\ref{fig:acc_similarity} shows each model’s accuracy across these bins. MedCLIP, PMC-CLIP, and LLaVA-Med achieve roughly 50\% accuracy in all bins. M3AE reaches 66\% accuracy in the lowest similarity bin, which decreases steadily as the similarity increases. This trend is expected. When replacement terms are highly dissimilar, models can readily distinguish the correct caption from the negative one, but discrimination becomes more difficult as term similarity increases.

\noindent\textbf{Summary of findings.} These results confirm—and indeed strengthen—our observation from Section~\ref{sec:evaluation_overall} and~\ref{sec:evaluation_mllm} that current multimodal models do not fully grasp medical concepts across vision and language. Our baseline observations remain robust under variations in caption length, vocabulary choice, and replacement-term similarity.
\section{Conclusion}
\label{sec:conclusion}
In this study, we proposed Medical-Checklist, a new benchmark designed to evaluate whether medical multimodal models possess the fundamental medical knowledge required to understand medical images. Unlike existing medical VQA datasets, which typically include clinically plausible distractors, our benchmark presents each medical image with a pair of captions: one correct and one clearly incorrect, generated by randomly substituting a medical term in the sentence. This setup enables us to assess whether models can reject medically nonsensical descriptions, thereby testing their grasp of basic medical concepts.

We evaluated four state-of-the-art medical multimodal models that have demonstrated strong performance on existing benchmarks. However, none of the models were able to pass the Medical-Checklist; they consistently failed to distinguish correct captions from clearly incorrect ones with the expected level of accuracy. These results reveal a significant gap between high performance on conventional benchmarks and genuine conceptual understanding, pointing to the limitations of current evaluation methodologies. This finding highlights the need for more rigorous and conceptually grounded evaluation methods—an essential step toward ensuring that multimodal AI systems can be deployed safely and reliably in real-world medical settings.

% ****************** Figure ******************
% ****************** Figure ******************

% ****************** Figure ******************
% ****************** Figure ******************

%
\newpage
\newpage
\section*{References}
\bibliographystyle{IEEEtran}
\bibliography{ref}

\newpage
\begin{IEEEbiography}[{\includegraphics[width=1in,height=1.25in,clip,keepaspectratio]{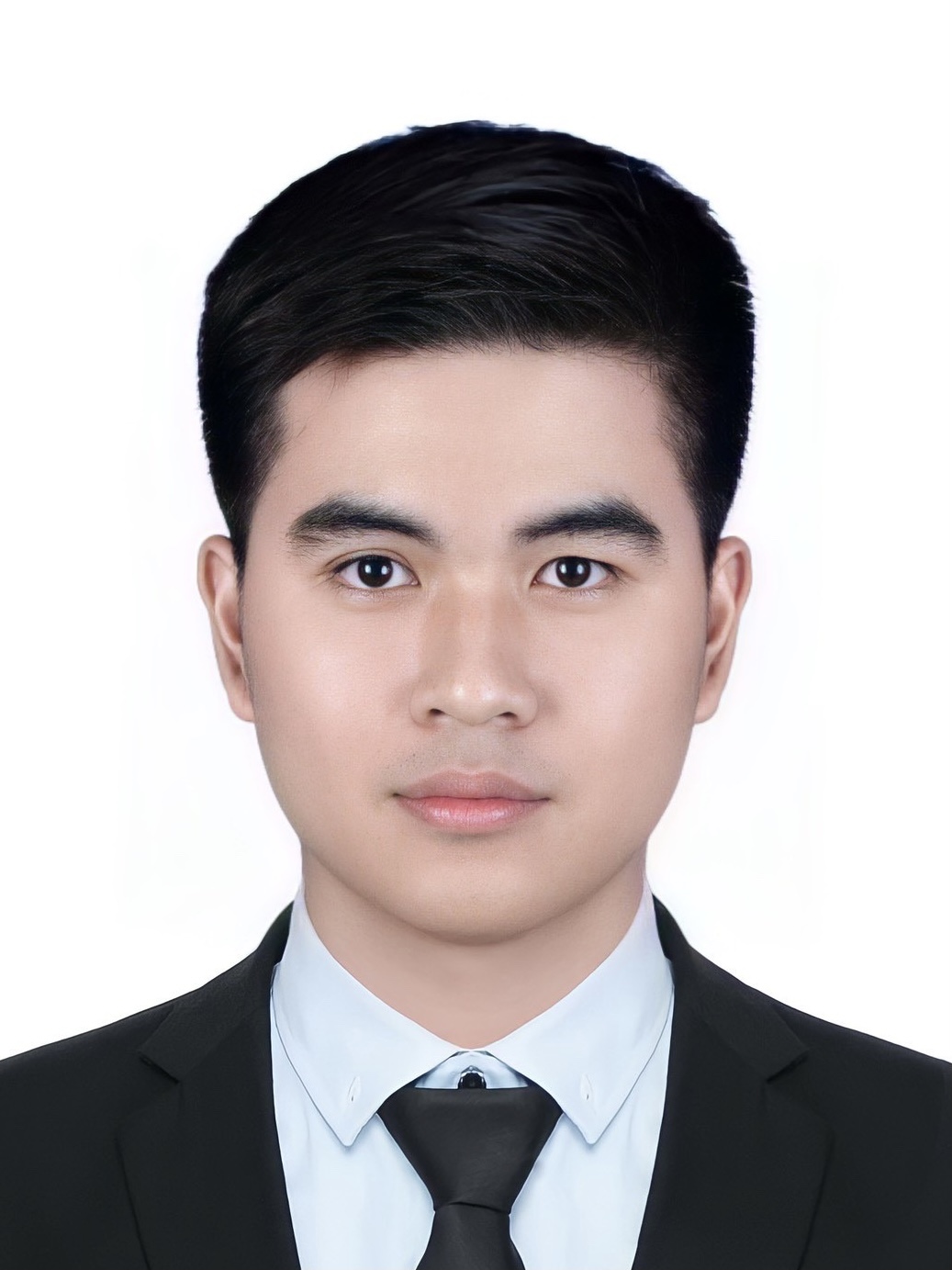}}]{Mr. Bannapol Limanond} is currently pursuing a Ph.D. degree at the Graduate School of Information Sciences, Tohoku University. His research interests are in the fields of medical computer vision and natural language processing.
\end{IEEEbiography}
\begin{IEEEbiography}
[{\includegraphics[width=1in,height=1.25in,clip,keepaspectratio]{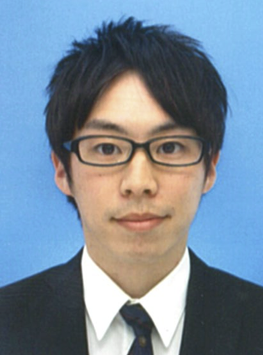}}]{Dr. Masanori Suganuma} received the Ph.D. degree from Graduate School of Environment and Information Sciences, Yokohama National University, in 2017. He is currently an Assistant Professor at Tohoku University. His research interests are in the field of computer vision and machine learning.
\end{IEEEbiography}
\begin{IEEEbiography}
[{\includegraphics[width=1in,height=1.25in,clip,keepaspectratio]{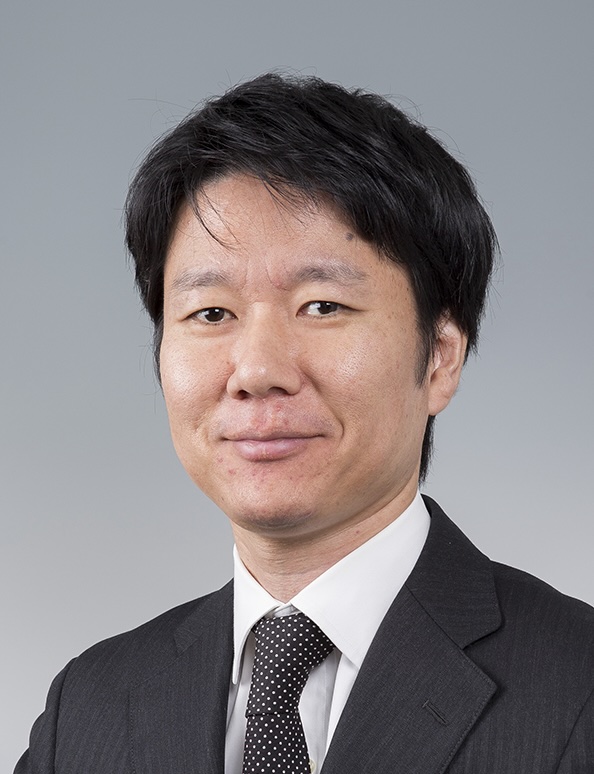}}]%
%[{\includegraphics[width=1in,height=1.25in,clip,keepaspectratio]{figures/author/suganuma_pic.png}}]
{Dr. Takayuki Okatani} earned his B.Eng., M.Sc., and Ph.D. degrees in Mathematical Engineering and Information Physics from the Graduate School of Engineering at the University of Tokyo in 1994, 1996, and 1999, respectively. He currently serves as a Professor in the area of computer vision at Tohoku University. In addition, he heads the Infrastructure Management Robotics Team at the RIKEN Center for Advanced Intelligence Project. With over 100 publications in peer-reviewed journals and conference proceedings, his work encompasses computer vision, deep learning, and multi-modal AI. He is an active member of several professional societies, including the IEEE Computer Society, the Information Processing Society of Japan, the Institute of Electronics, Information and Communication Engineers, and the Society of Instrument and Control Engineers.
\end{IEEEbiography}

\newpage

\end{document}